\documentclass{article}

\usepackage{PRIMEarxiv}

\usepackage[utf8]{inputenc} 
\usepackage[T1]{fontenc}    
\usepackage{hyperref}       
\usepackage{xurl}           
\usepackage{booktabs}       
\usepackage{amsfonts}       
\usepackage{nicefrac}       
\usepackage{microtype}      
\usepackage{fancyhdr}       
\usepackage{multirow}
\usepackage{orcidlink}
\usepackage{graphicx}       
\graphicspath{{figures/}}     

\title{BLDNet: A Semi-supervised Change Detection Building Damage Framework using Graph Convolutional Networks and Urban Domain Knowledge
\thanks{\textit{\underline{Citation}}: 
\textbf{Authors. Title. Pages.... DOI:000000/11111.}} 
}

\author{
  Ali Ismail\,\orcidlink{0000-0001-8636-1373} \\
  Department of Electrical and Computer Engineering \\
  American University of Beirut \\
  Beirut, Lebanon\\
  \texttt{ali.ai@live.com} \\
   \And
  Mariette Awad\,\orcidlink{0000-0002-4815-6894} \\
  Department of Electrical and Computer Engineering \\
  American University of Beirut \\
  Beirut, Lebanon\\
  \texttt{mariette.awad@aub.edu.lb} \\
}

\begin{document}
\maketitle

\begin{abstract}
Change detection is instrumental to localize damage and understand destruction in disaster informatics. While convolutional neural networks are at the core of recent change detection solutions, we present in this work, BLDNet, a novel graph formulation for building damage change detection and enable learning relationships and representations from both local patterns and non-stationary neighborhoods. More specifically, we use graph convolutional networks to efficiently learn these features in a semi-supervised framework with few annotated data. Additionally, BLDNet formulation allows for the injection of additional contextual building meta-features. We train and benchmark on the xBD dataset to validate the effectiveness of our approach. We also demonstrate on urban data from the 2020 Beirut Port Explosion that performance is improved by incorporating domain knowledge building meta-features.
\end{abstract}

\keywords{Change detection \and Building damage \and Urban data science \and Disaster informatics \and Graph convolutional networks \and Semi-supervised learning \and Meta-features \and Siamese networks \and Domain knowledge}

\section{Introduction}
Change detection (CD) is a sub-field of remote sensing concerned with identifying and localizing differences in surface objects using images taken at different times. CD has been successfully applied in many different fields such as forest monitoring \cite{khan_forest_2017,tan_improved_2013,woodcock_monitoring_2001}, surface water monitoring \cite{rokni_new_2015,song_change_2019}, sea ice monitoring \cite{gao_sea_2019,gao_transferred_2019}, landslide monitoring \cite{ding_automatic_2016}, tsunamis \cite{bovolo_split-based_2007,sublime_automatic_2019}, fires \cite{fraser_multitemporal_2003}, urban land use \cite{zhang_detecting_2019,cao_land-use_2019,lu_detection_2010} and planetary surfaces \cite{kerner_toward_2019}. In disaster informatics such as the world trade center attacks and hurricane Katrina, \cite{robila_use_2006} building damage assessment was mostly emphasized \cite{gupta_xbd_2019}. 

Traditionally, CD was performed using classical image processing techniques \cite{asokan_change_2019}. With the increasing availability of huge amounts of data and with the AI revolution, new CD techniques based on machine learning have emerged \cite{shi_change_2020}. CNNs are currently the most used architecture for CD \cite{khelifi_deep_2020}. Various CNN architectures have been used for CD such as object detection \cite{khan_forest_2017} and semantic segmentation \cite{peng_end--end_2019}. Since most CD applications use pre and post images, Siamese CNN architectures are particularly popular \cite{zhan_change_2017,daudt_urban_2018}. Due to the temporal aspect of CD, recurrent neural networks have also been used where the pre image is considered to be the first timestep followed by the post image \cite{lyu_learning_2016}. These models have been subsequently merged with CNNs and Siamese CNNs to exploit their powerful image representation abilities \cite{mou_learning_2019,chen_change_2020,jing_object-based_2020}. More specifically, CD for building damage has been addressed using similar approaches \cite{ji_building_2019,nex_structural_2019,jiang_pga-siamnet_2020,kalantar_assessment_2020,miura_deep_2020,wheeler_deep_2020,sublime_automatic_2019}.

However, CNNs can only extract spatial information within the defined neighborhood of the convolution kernel. Graph convolutional networks (GCN) instead model data as graphs \cite{kipf_semi-supervised_2017} and explore non-euclidean relationships that go beyond the pixel neighborhood defined by a CNN kernel \cite{hong_graph_2020}. GCNs have outperformed CNN-based models in many computer vision applications \cite{chaudhuri_siamese_2019,hong_graph_2020} including urban change detection \cite{saha_semisupervised_2020}.

On disaster onset, a lot of data is available almost immediately \cite{robila_use_2006}. While such data will enable training connectionist models, labeling is a major challenge for such humanitarian applications.  
Besides, many cities include a mixture of buildings of various shapes, heights, time periods, building material and architectural styles. Usually, these buildings compete for the urban footprint and are in close proximity to each other. These architectural and structural differences can cause buildings to interact differently with the disaster and therefore sustain damage to varying levels. While current approaches rely strictly on satellite or aerial imagery, it may not be possible to predict well the damage level without the aforementioned architectural differences embedded in the urban data.


Based on these gaps found in state-of-the-art research, we cast building damage CD as a node classification problem. This formulation leverages both the local and neighborhood features and patterns in the data. We present BLDNet, a hybrid CNN and GCN novel architecture that is trained in a semi-supervised manner to obtain predictions in a timely manner. Furthermore, we demonstrate how architectural and contextual building features which we henceforth call meta-features can be incorporated as domain knowledge into the graph to improve the prediction results.

The contributions of this paper are:
\begin{itemize}
    \item A novel formulation of building damage identification as a graph node classification to learn representations based on local features as well as relationships with neighboring samples.
    \item BLDNet, a novel architecture based on a Siamese CNN combined with a GCN trained in a semi-supervised manner to reduce the number of labeled samples needed to obtain new predictions.
    \item A case study on the urban data collected from Beirut Port 2020 explosion with  knowledge domain injection.
\end{itemize}

To the best of our knowledge, there is no prior work which fuses both images and meta-features in urban data for CD of building damage. Also, there is no prior work which adopts a meta-feature framework with GCN.

The rest of this document is organized as follows. In Section \ref{sec:litt}, we review related CD work and similar GCN applications. In Section \ref{sec:methodology}, we explain BLDNet methodology. Section \ref{sec:experiments} details the datasets used, experiments performed, and results interpretation while Section \ref{sec:conclusion} concludes the work with future research directions.

\section{Related Work}\label{sec:litt}
In this section we review deep learning approaches for building damage detection as well as approaches for accelerating the process of obtaining new detections. For a more comprehensive review of CD and AI, the reader is referred to the surveys by Shi et al. \cite{shi_change_2020} and Khelifi and Mignotte \cite{khelifi_deep_2020}.

\subsection{AI for Building Damage Change Detection}
The objective of CD in building damage is to localize damage and assess its severity. An object detection problem using Faster Region-based CNN \cite{wang_change_2018}, a classification using single stream Siamese CNNs \cite{kalantar_assessment_2020,wheeler_deep_2020} and a semantic segmentation using pyramid encoder-decoders \cite{jiang_pga-siamnet_2020} which often have Siamese backbones \cite{weber_building_2020} were proposed. Because building damage data suffers from severe class imbalance, some works adopted weighted loss functions \cite{su_technical_2020,wheeler_deep_2020} or generated positive samples using pixel level reduction and shifting \cite{ji_building_2019} or generative adversarial networks \cite{su_technical_2020}.

\subsection{Methodologies for Fast Change Detection}
To accelerate damage detection on new disasters, literature relies either on transfer from one disaster/region to another which allows inference without training or on semi-supervised learning which enables training a model with fewer labeled samples.

\subsubsection{Cross-Domain Transfer for Disaster Damage Detection}
The works  \cite{nex_structural_2019,xu_building_2019} concerned with developing methodologies to infer new disasters used model architectures similar to the ones described in the previous paragraph. Other studies extrapolated to different disasters types \cite{miura_deep_2020,benson_assessing_2020,bai_pyramid_2020,yang_transferability_2021,zheng_building_2021}. In many of these cases, transferability is dependent on the similarity between to the train and test regions \cite{nex_structural_2019} and is usually degraded by data heterogeneity \cite{nex_structural_2019,yang_transferability_2021}. Other solutions proposed including a small number of test samples with the training data \cite{xu_building_2019,yang_transferability_2021}, a multi-domain adaptive batch normalization and a stochastic weight averaging \cite{benson_assessing_2020}.

\subsubsection{Semi-supervised Learning for Change Detection}
Pati et al. automatically labeled data using clustering and fuzzy logic and used the labels to train a supervised model \cite{pati_novel_2020}. Other works detected building damage with an outlier detection deep autoencoder \cite{sublime_automatic_2019} and an anomaly detecting generative adversarial network \cite{tilon_post-disaster_2020}. Peng et al. proposed a semi-supervised generative adversarial network where the generator is a supervised segmentation network. The discriminator was trained to distinguish between the generator's prediction on unlabeled samples and the ground truth of labeled samples to improve the unlabeled predictions \cite{peng_semicdnet_2021}.

\subsection{Graph Convolutional Networks in Computer Vision}
The representation of images as graphs enabled GCNs to solve many computer vision problems.
In the context of remote sensing image retrieval, a Siamese GCN was built using pre-computed image segment features as nodes. Adjacent nodes were connected with the edge weight being the pairwise centroid pixel and segment orientation difference \cite{chaudhuri_siamese_2019,khan_graph_2019}. To improve multilabel image classification, Chen et al. mapped the logical relationships between the different classes in a directed graph and trained a stack of GCNs combined with the output of a CNN \cite{chen_multi-label_2019}. \cite{mou_nonlocal_2020} exploited the semi-supervised nature of GCNs and the ability to learn from image-wide relationships using non-local graph representation for land use classification. They built their graph using each pixel as a node and the similarity between every pair of nodes was used to build the edges to indicate the likelihood of two nodes belonging to the same class. Hong et al. similarly built a supervised GCN variant  trained using mini-batches for hyperspectral image classification and tested it combined with a GCN on different output fusion schemes \cite{hong_graph_2020}.

\subsection{State-of-the-Art Work}
\begin{figure}[htb]
    \centering
    \includegraphics[scale=0.55]{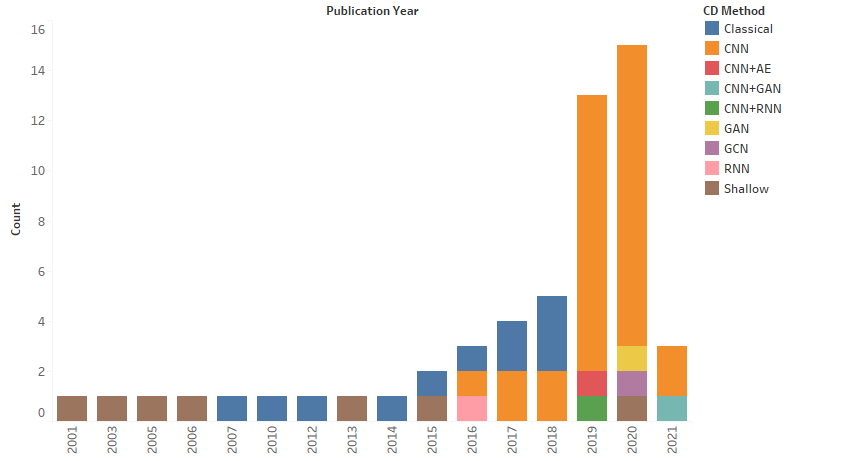}
    \caption{Evolution of CD methods over the years.}
    \label{fig:litt_method}
\end{figure}

In recent years, deep learning has been the most adopted approach for CD, owing to its success in computer vision. We show in Figure \ref{fig:litt_method} a timeline for CD approaches over the years. Classical and shallow learning approaches have seen a decrease in use since 2019. This was accompanied by a sustained sharp increase in CNN adoption. Moreover, additional architectures have appeared in 2020 and 2021 such as hybrid GAN-based models as well as GCN.

Differently from existing similar works, we formulate a graph node classification framework for building disaster damage detection. In \cite{saha_semisupervised_2020}, the graph nodes were constructed using pre and post superpixel parcels to compute summary node features. In our work, we learn deep pixel feature maps as node features instead of calculating summary features as it is done in traditional CNN frameworks. For graph edges, Saha et al. \cite{saha_semisupervised_2020} connected only parcels with an adjacent pixel while we use a geometric triangulation to ensure that every building is connected to all of its surrounding neighbors. Furthermore, we embed meta-features into our graph to improve learning based on domain knowledge which is different from aforementioned works that relied on image data only. In terms of accelerating damage assessment for new events, the bulk of prior art focuses on pre-trained models for direct inference. However, the generalization of these models is effective to varying degrees. It often hinges on the similarity between the test and train data and degrades with data heterogeneity \cite{yang_transferability_2021}. Many works even included a few labeled samples from the test set with the training set which makes the process analogous to semi-supervised learning \cite{xu_building_2019,yang_transferability_2021}. Because semi-supervised models are easier to develop, we capitalized on semi-supervised learning for efficient deployment. Table \ref{tab:sota} highlights the differences between our work and most related prior art.

\begin{table}[htb]
\centering
\caption{Qualitative Comparison with the State-of-the-art.}
\label{tab:sota}
\resizebox{\textwidth}{!}{%
\begin{tabular}{@{}cccccc@{}}
\toprule
\textbf{Work} & \textbf{Building Damage} & \textbf{Uses GCN} & \textbf{Semi-supervised} & \textbf{Domain Transfer} & \textbf{Uses Meta-features} \\ \midrule
\textbf{\cite{yang_transferability_2021}} & Yes & No & No & Yes & No \\
\textbf{\cite{saha_semisupervised_2020}} & No & Yes & Yes & No & No \\
\textit{\textbf{This Work}} & Yes & Yes & Yes & No & Yes \\ \bottomrule
\end{tabular}%
}
\end{table}

\section{Methodology}\label{sec:methodology}
In this section, we present details about the BLDNet formulation and its training configurations.

\subsection{Graph Data Formulation}
BLDNet is a graph-based approach that exploits both local image features as well as relationships with neighboring samples. It assumes the existence of building footprint polygons which are available in the xBD dataset and the Beirut Dataset from the Beirut Recovery Map\footnote{https://openmaplebanon.org/beirut-recovery-map}. In case these footprints were not available, they can be obtained using a building footprint detector \cite{wheeler_deep_2020}.

\begin{figure}[htb]
    \centering
    \includegraphics[scale=0.5]{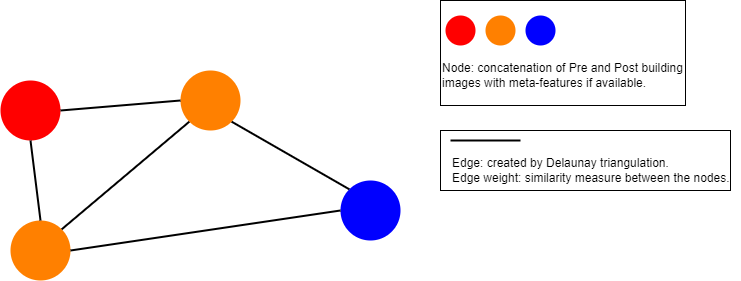}
    \caption{Concept of the Graph Formulation. Different node colors indicate nodes' different classes.}
    \label{fig:graph_form}
\end{figure}

\begin{figure}[htb]
    \centering
    \includegraphics[scale=0.15]{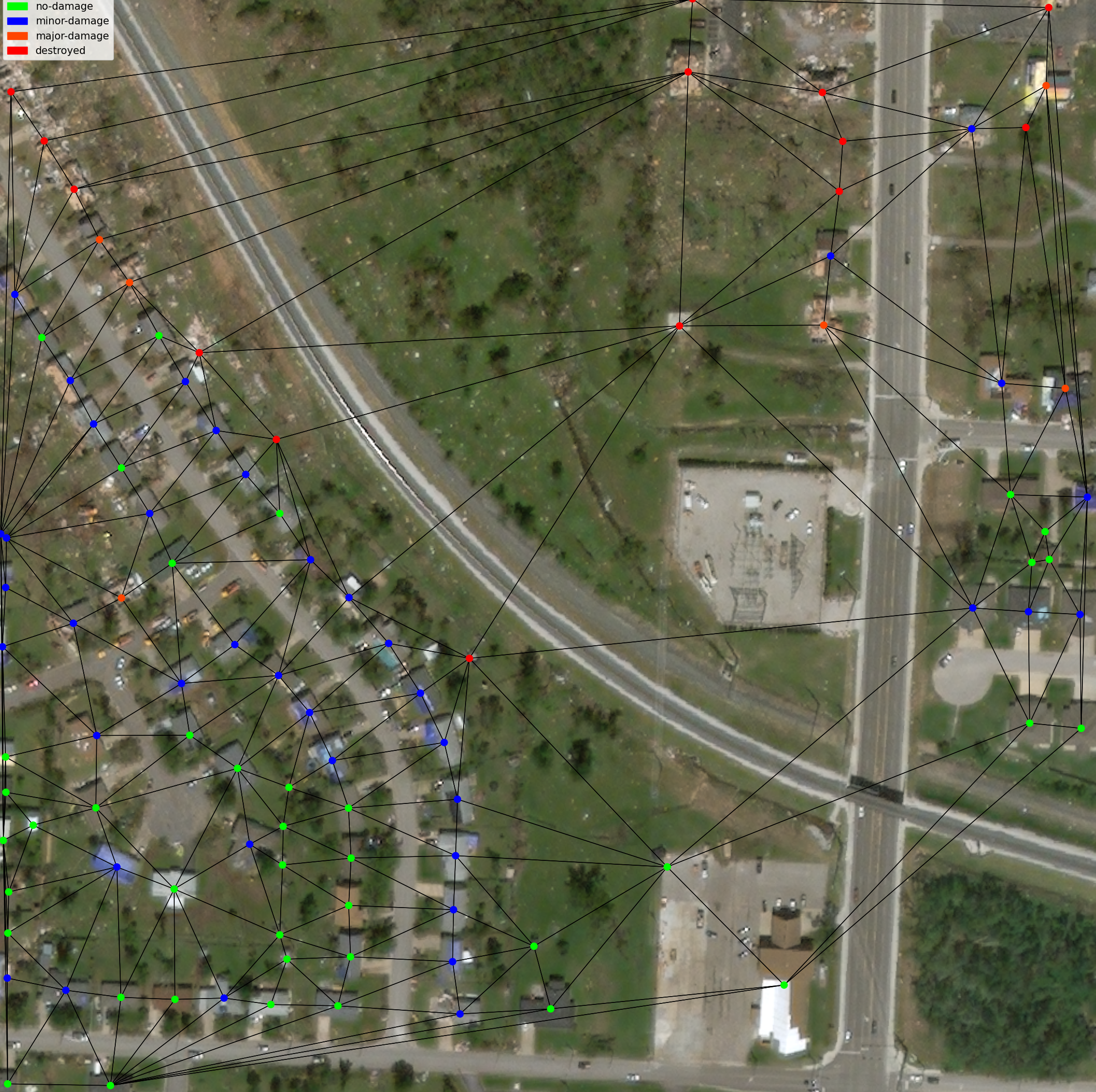}
    \caption{Subgraph implemented on a sample from the Joplin Tornado set.}
    \label{fig:graph_joplin}
\end{figure}

Our formulation builds an undirected acyclic graph using the building polygons. Each node represents one building. The node features are the concatenation of the pre and post image crops defined by the rectangular envelope of the building polygons. The crops are resized to a width and height of $128 \times 128$ in order to unify their size and reduce the memory footprint of the graph. The resulting feature vector is of size $N = 128 \times 128 \times 3 \times 2 = 98304$. To exploit domain knowledge information, we embed the meta-features at the node level which increases the dimension of the vector to $N = 98324$. The edges are constructed using a Delaunay triangulation \cite{lee_two_1980} based on the UTM (Universal Transverse Mercator) building envelope centroid coordinates. Each edge is given a weight measuring the similarity between the connected nodes according to the equation used by Saha et al. \cite{saha_semisupervised_2020} to build their adjacency matrix. Figure \ref{fig:graph_form} shows a conceptual sketch of the proposed graph formulation while Figure \ref{fig:graph_joplin} shows a realized implementation of a subgraph from the xBD Joplin tornado.

The semi-supervised GCN requires the entirety of the data during training to be built into a single graph. Only a relatively small number of nodes is labeled during training and the model learns to propagate labels along the graph edges to the rest of nodes. We thus constructed one large graph out of each selected region.

\subsection{Model Architecture}
Using our graph formulation, we employ graph convolutions to aggregate information from neighboring nodes. However, given that preliminary experiments have also shown that it is important to have a reliable way of extracting local features, we developed BLDNet, a hybrid architecture combining a CNN and a GCN. The CNN backbone is used to extract local image features and the GCN operator is used to aggregate them with CNN features of connected nodes. The CNN backbone is a Siamese ResNet34 network with the classification layers removed \cite{he2015deep}. The output of the two-stream network is the difference between the individual stream outputs which is piped into a GCN \cite{kipf_semi-supervised_2017} whose output is the classification. The ReLu activation function is used for all layers except the last layer which uses a Softmax activation. The ResNet weights are initialized with the ImageNet weights and trained along with the entire model.

\begin{figure}[htb]
    \centering
    \includegraphics[width=\textwidth]{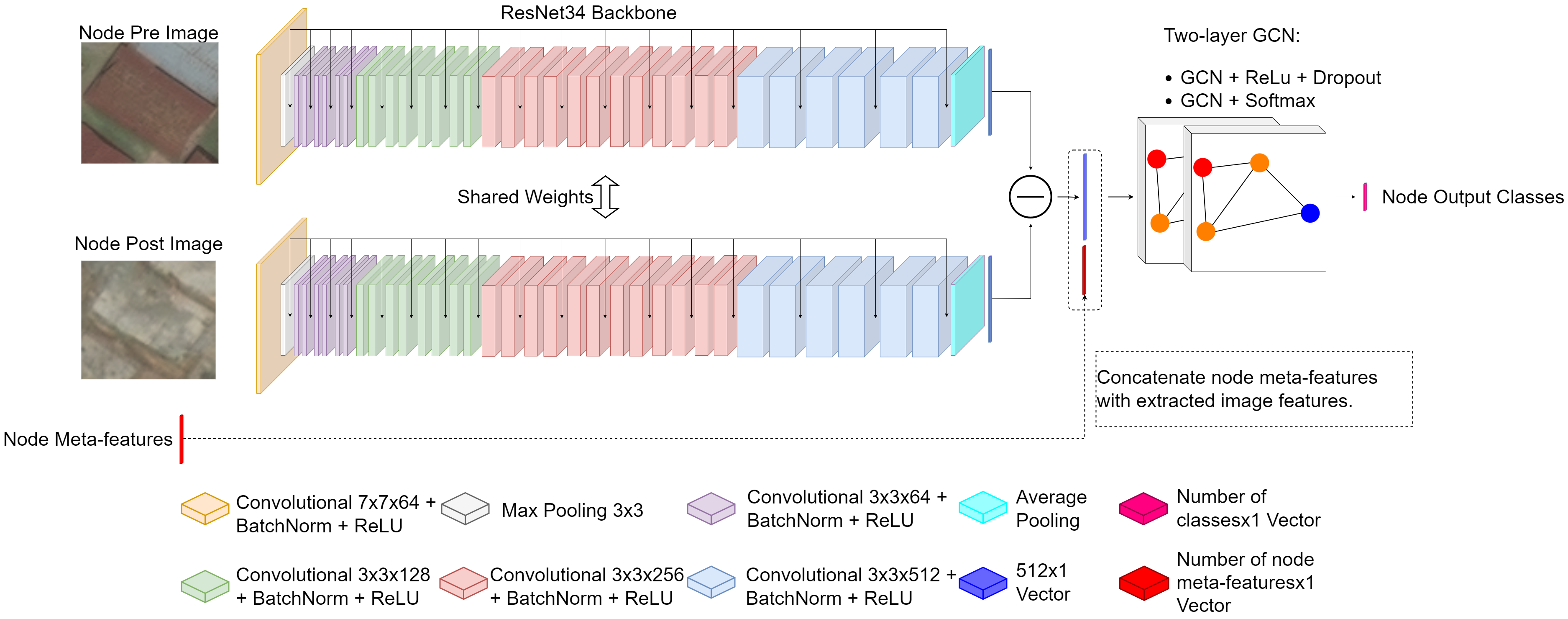}
    \caption{Architecture of BLDNet. Objects in dotted lines indicate meta-feature injection if these features are available.}
    \label{fig:model_arch}
\end{figure}

 The meta-features are concatenated with the output of the ResNet backbone. The resulting vector is then inputted through the GCN layers. While this meta-feature injection method was adopted in other works, it was for different application areas than urban damage assessment  \cite{calderisi_improve_2019,ellen_improving_2019}. The diagram in Figure \ref{fig:model_arch} shows the overall layout of BLDNet.

\section{Experiments and Results}\label{sec:experiments}
In this section, we present the data that was used along with the data preprocessing steps. We describe the experiments and procedures to implement and evaluate BLDNet. We also analyze and discuss our results.

\subsection{Datasets}
The proposed approach was tested on both a public dataset to position it with respect to the literature as well as the Beirut Port Explosion dataset to demonstrate the effectiveness of our novel meta-feature injection approach.

\subsubsection{xBD Dataset}
The xBD dataset has different cases of city damage for different disaster types \cite{gupta_xbd_2019}. It contains a collection of manually annotated pre and post Worldview 2 images. The dataset is organized into two big sets. The Tier 1 set contains a set of disasters subdivided into train, test and hold partitions. The Tier 3 set introduces additional disasters.

xBD is the most diverse in terms of regions, disaster types, buildings and urban density. It contains image chips of earthquake, tsunami, flood, volcano, wildfire and tornado/hurricane disasters across sixteen regions. The landscape in these different regions varies from urban scenes such as Mexico to largely rural regions such as Guatemala.

\begin{figure}[htb]
    \centering
    \includegraphics[scale=0.55]{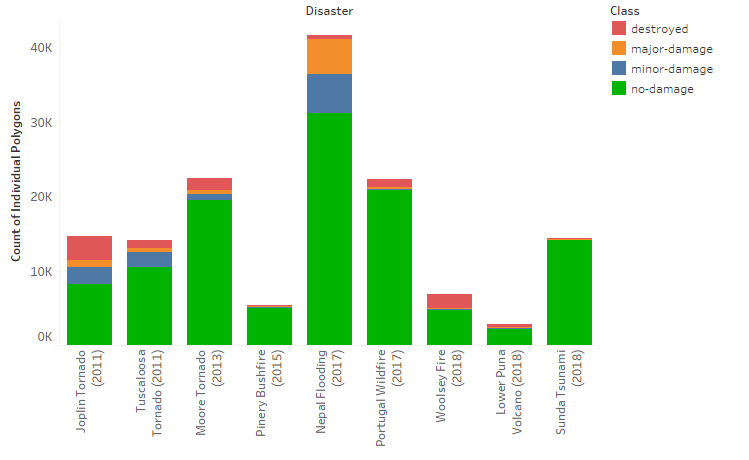}
    \caption{xBD Tier 3 number of buildings per class.}
    \label{fig:tier3_poly}
\end{figure}

\begin{figure}[htb]
    \centering
    \includegraphics[scale=0.55]{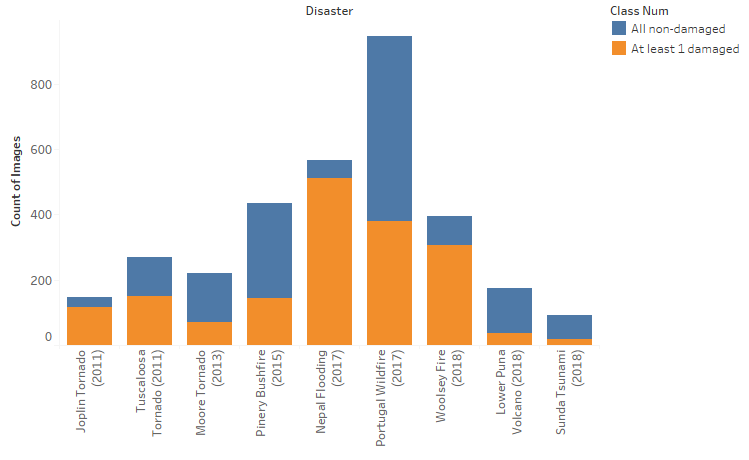}
    \caption{xBD Tier 3 image chips with at least one damaged building.}
    \label{fig:tier3_zones}
\end{figure}

The dataset is annotated according to a four-class joint damage scale: no damage, minor damage, major damage and destroyed. It suffers from severe class imbalance as the majority of buildings are non-damaged. On average, $80.4\%$ of individual buildings across the Tier 3 regions are non-damaged (Figure \ref{fig:tier3_poly}). Also, since each region is divided into equal image chips, we investigated the number of chips with at least one damaged instance and found that about $50\%$ on average do not contain any damaged buildings (Figure \ref{fig:tier3_zones}). These insights helped shape our data sampling strategy concerning which samples can be discarded.

\subsubsection{Beirut Explosion Dataset}
Beirut as a city, is an architectural mosaic with irregular and heterogeneous building patterns with many different layers of architectural styles gradually added over decades. The architecture spans many centuries and civilizations as it is home to Roman heritage sites and civil war (1975-1990) ruins. However, these structures are not located far from luxurious commercial and financial districts and modern residential high-rises. The architectural styles that dot the city are also varied and include Ottoman, European and Gothic patterns \cite{bucknell_2018}.

The unfortunate Beirut port explosion occurred on August 4, 2020 resulting in the death of nearly 135 people, around 5000 injuries and causing severe structural damage to the surrounding area leaving more than 300000 people homeless \cite{reliefweb}. The strength of the blast was estimated to be equivalent to the twentieth of the Hiroshima bomb and was listed as the strongest non-nuclear explosion in the 21st century \cite{rigby_preliminary_2020}.
Worldview 2 satellite products covering a large portion of Beirut were donated by Maxar Technologies. The first view was taken on July 31 2020 and the second post disaster on August 5 2020. The damage classification was provided by the Open Map Lebanon team\footnote{\url{https://openmaplebanon.org/beirut-recovery-map}}. These annotations use a four class damage scale: minor, moderate, major and severe damage. Furthermore, the heterogeneous and irregular urban and architectural landscape of Beirut motivated us to use contextual meta-features. These enable our model to learn the patterns and interactions resulting from buildings' non-uniformity. This dataset was obtained from Krayem et al. \cite{krayem_machine_2021}.

\begin{figure}[htb]
    \centering
    \includegraphics[scale=0.4]{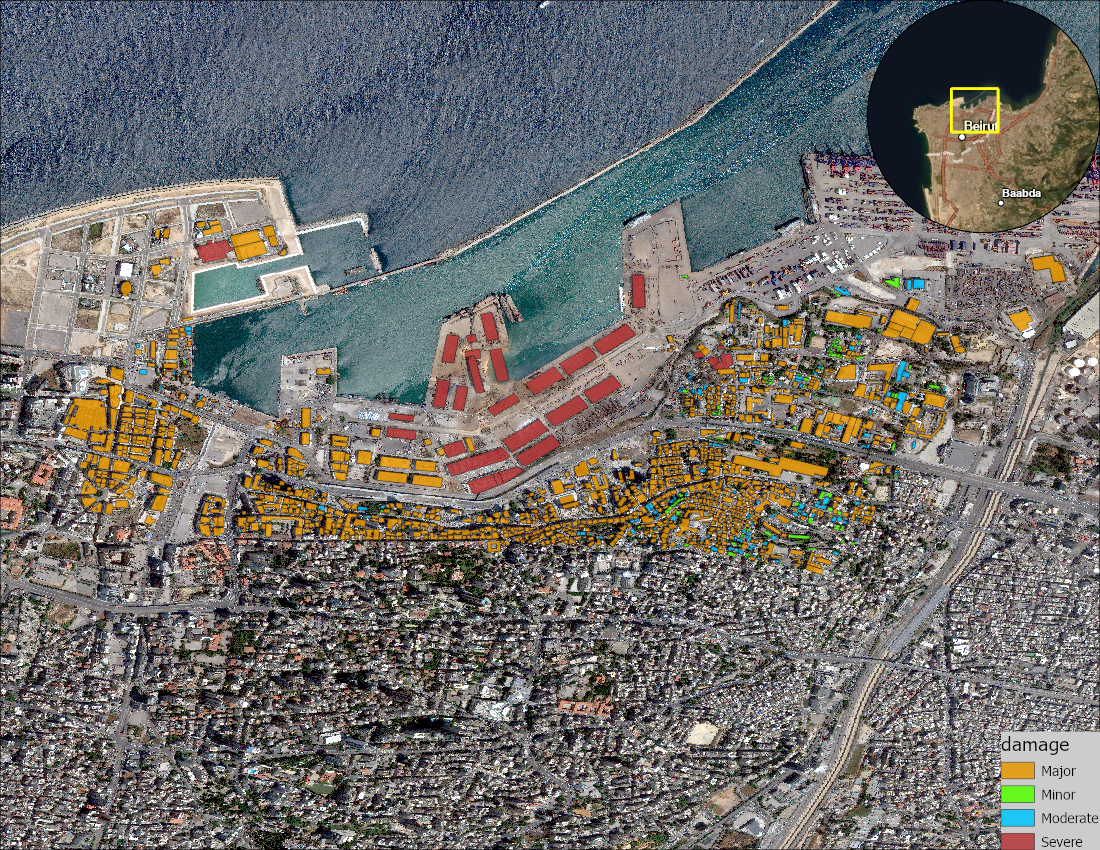}
    \caption{Overview of the selected area.}
    \label{fig:beirut_area}
\end{figure}

\begin{figure}[htb]
    \centering
    \includegraphics[scale=0.3]{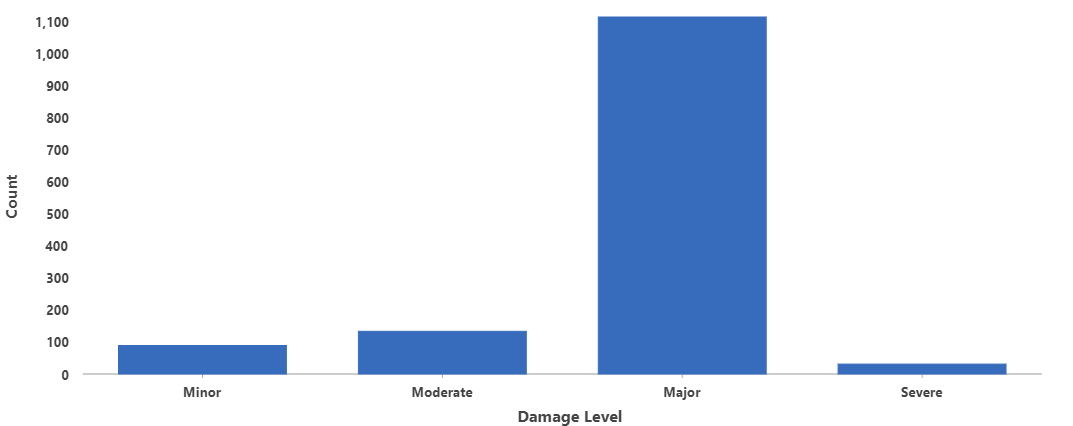}
    \caption{Class distribution of the selected samples.}
    \label{fig:beirut_dist}
\end{figure}

The two satellite products (pre and post) were used to produce pansharpened color images. The building polygons were overlaid on top of the images which were georeferenced to fit the polygons. A subset of buildings was selected due to memory constraints. A view of this area as well as the class distribution of the included polygons are shown in Figure \ref{fig:beirut_area} and Figure \ref{fig:beirut_dist} respectively.

\begin{figure}[htb]
    \centering
    \includegraphics[scale=0.4]{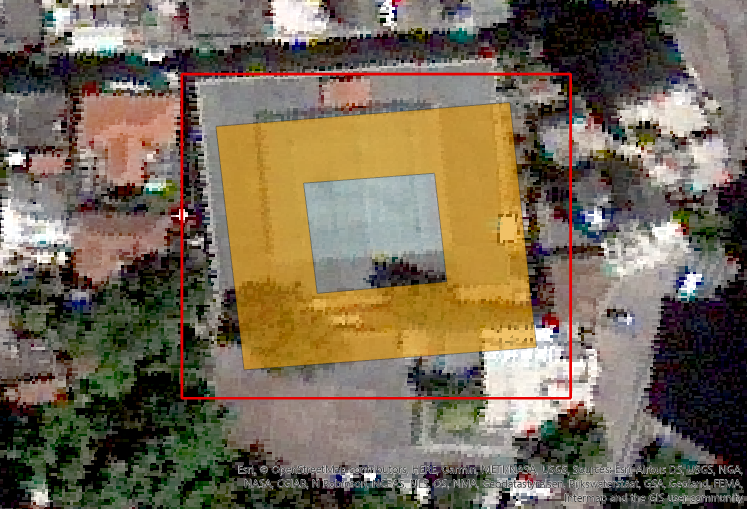}
    \caption{Buffered mask (red) around the original polygon (orange).}
    \label{fig:beirut_buffered_mask}
\end{figure}

Finally, we created a rectangular envelope around each polygon buffered by five meters in each direction (Figure \ref{fig:beirut_buffered_mask}) because the polygons were drawn to the shape of the building rooftops. For tall structures, this would have resulted in the cropped image showing only a small portion of the visible part of the building. Buffering the polygons also helps with mitigating the effect of georeferencing error. Other works encountered such difficulties and adopted similar solutions \cite{zheng_building_2021,miura_deep_2020}.

\begin{figure}[htb]
    \centering
    \includegraphics[scale=0.5]{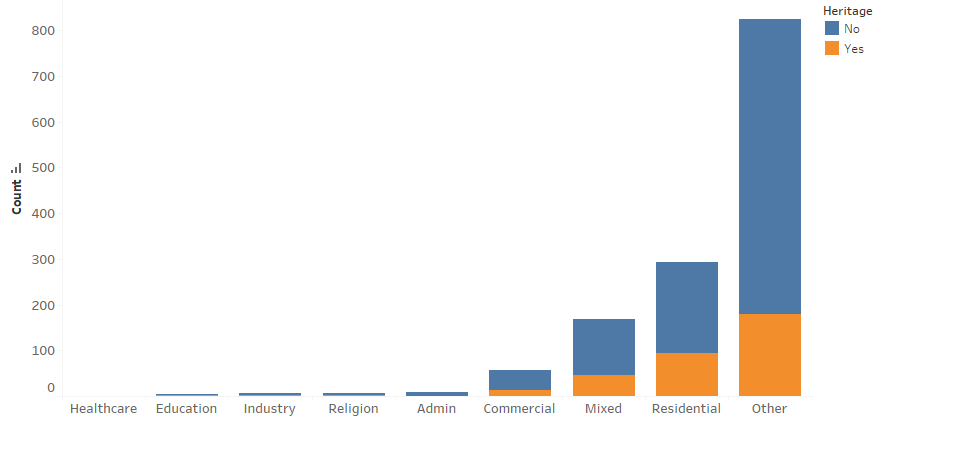}
    \caption{Counts of Beirut heritage buildings and Beirut buildings by function.}
    \label{fig:beirut_heritage_function}
\end{figure}

The meta-features dataset contains multiple building features collected from multiple sources. We retained features related to the number of apartments, area, perimeter, number of floors, digital surface model (DSM), building height, construction year, construction era, heritage status and building function. Despite consolidating the different columns containing the same feature coming from different sources to minimize missing values, the data still had a lot of null entries such as the value "Other" in Figure \ref{fig:beirut_heritage_function}. This Figure shows the distributions of the Heritage and Building Function attributes while Table \ref{tab:beirut-meta-stats} reports on some statistics including the number of missing values for each attribute. Finally, all columns were normalized to a range between zero and one, with zero representing missing values.

\begin{table}[htb]
\centering
\caption{Summary Statistics of the Beirut Meta-data.}
\label{tab:beirut-meta-stats}
\resizebox{\textwidth}{!}{%
\begin{tabular}{|c|c|c|c|c|c|c|c|c|}
\hline
 & \textbf{Number of apartments} & \textbf{Mean DSM} & \textbf{Mean building height} & \textbf{Area} & \textbf{Perimeter} & \textbf{Era} & \textbf{Built year} & \textbf{Floors} \\ \hline
\textbf{Count} & 535 & 1023 & 1023 & 1257 & 1257 & 711 & 789 & 870 \\ \hline
\textbf{Number of nulls} & 831 & 343 & 343 & 109 & 109 & 655 & 577 & 496 \\ \hline
\textbf{Mean} & 6.72 & 41.24 & 18.47 & 568.03 & 86.95 & 2.74 & 1954.41 & 4.17 \\ \hline
\textbf{Std} & 6.95 & 18.39 & 10.88 & 1484.34 & 76.44 & 0.97 & 36.93 & 2.63 \\ \hline
\textbf{Min} & 1 & 3.78 & 1.66 & 17.58 & 17.81 & 1 & 1219 & 1 \\ \hline
\textbf{Max} & 60 & 113.70 & 95.65 & 16574.86 & 1062.72 & 5 & 2021 & 26 \\ \hline
\end{tabular}%
}
\end{table}

\subsection{Experimental Setup}
All of the experiments were implemented using Python 3.8, Pytorch 1.7.1 and Pytorch Geometric 1.7.0 \cite{fey_fast_2019}. The code base that was used to perform these experiments is available\footnote{\url{https://gitlab.com/awadailab/gcn-remote-sensing}}. The Adam optimizer \cite{kingma2017adam} was used with the categorical crossentropy loss function which included class weights to mitigate the impact of class imbalance \cite{wheeler_deep_2020}. Additionally, the "major damage" and "destroyed" classes in the xBD dataset were merged into a single class because they are hard to distinguish. This aggregation was adopted by Xu et al. on their dataset \cite{xu_building_2019}. For similar reasons, we merge the "minor" and "moderate" classes in the Beirut dataset. For model evaluation and comparison, we compute the accuracy, precision, recall, specificity and the F1 score.


The number of neurons, number of layers, dropout rate and learning rate were tuned to reduce overfitting and improve performance on the test set. The final adopted configuration is two graph convolutional layers with 32 neurons and a dropout rate of 0.5. The learning rate was set empirically to 0.0003. All of our BLDNet experiments that included preprocessing the data, building the graph, training and producing predictions took around 10 to 15 minutes to complete on a virtual machine running on 8 cores of an AMD EPYC 7551 32-Core Processor with an Nvidia V100 32GB GPU.

\subsection{Semi-supervised GCN with the xBD dataset}
To evaluate the merit of BLDNet we train and test on each of the Pinery Bushfire, Joplin Tornado and Nepal Flooding events from the xBD Tier 3 set. This experiment requires all of the data during training because it is semi-supervised. Moreover, in the case of GCN, all of the data needs to be loaded at once into memory to build the full graph.

\begin{figure}[htb]
    \centering
    \includegraphics[scale=0.6]{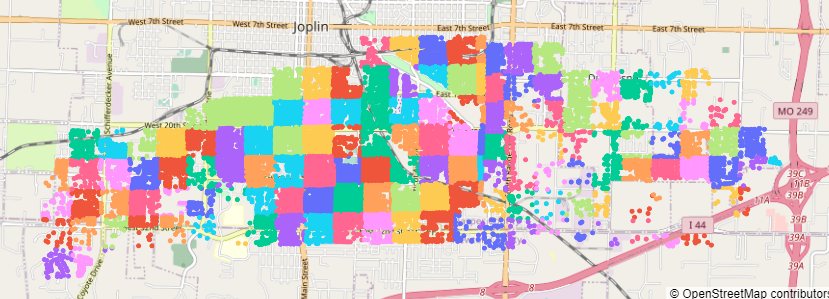}
    \caption{Image chips of Joplin Tornado as squares of different color.}
    \label{fig:joplin_zones}
\end{figure}

To efficiently manage memory resources, we pruned the unlabeled samples. As previously mentioned, each disaster event is divided into equal image chips (Figure \ref{fig:joplin_zones}). We have also shown (Figure \ref{fig:tier3_zones}) that many chips do not have any damaged samples at all because the damage is more concentrated around the damaging force.

Therefore, we discarded image chips that do not have any damaged samples. If the data still needed to be reduced, a random subset is sampled in way to preserve the class distribution of the original data.

The obtained samples were fed as input to build the graph. Each node in the graph is marked as being either training, testing or hold. This assignment is done randomly while maintaining the same class distribution. The training nodes are a small number of nodes for which the loss function will be computed during training to optimize the model. Performance metrics were computed on testing samples after every epoch for model selection. Once training, the metrics are recorded on the hold set as well as the entire graph.

\begin{figure}[htb]
    \centering
    \includegraphics[scale=0.5]{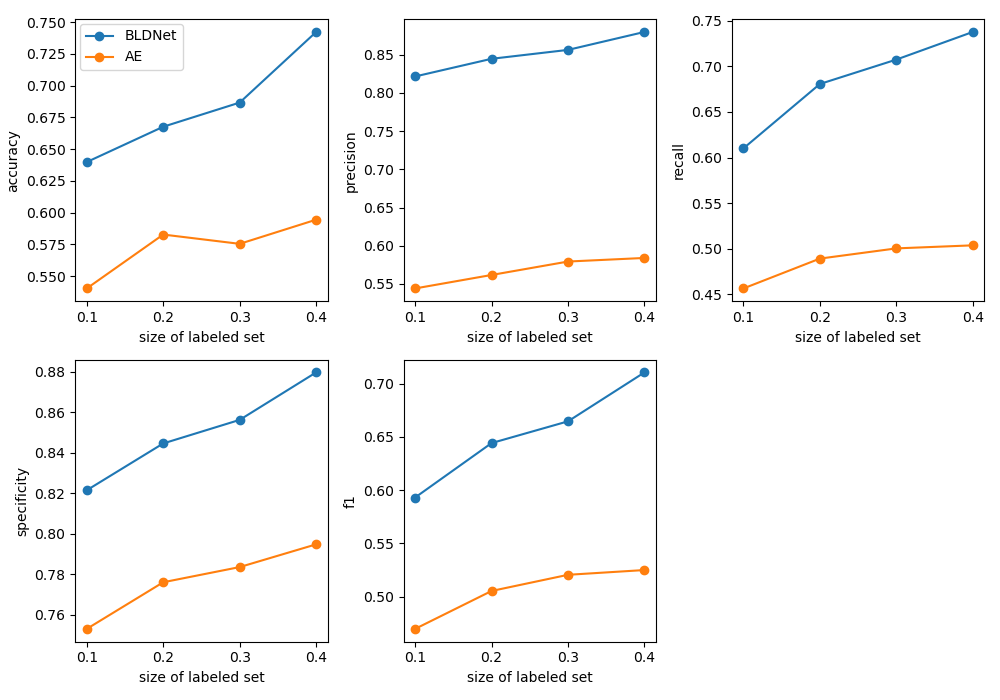}
    \caption{Model performance with respect to the training data size.}
    \label{fig:sensitivity}
\end{figure}

We experimented with different sizes for the training set and found that using 20\% of the data to be the best compromise in terms of performance and number of training samples (Figure \ref{fig:sensitivity}).

\subsubsection{Benchmarking with Multiresolution Autoencoder}
To benchmark our model, we trained the semi-supervised Multiresolution Autoencoder \cite{ienco_semi-supervised_2018} on the same data that was used to train BLDNet with exactly the same train, test and hold splits. This model was also used for benchmarking by Saha et al. \cite{saha_semisupervised_2020}. Table \ref{tab:gcn_ae_xbd} shows the results for both BLDNet and the multiresolution autoencoder.

\begin{table}[htb]
\centering
\caption{Comparison between BLDNet and the Multiresolution Autoencoder (AE) with bold italics indicating better performance.}
\label{tab:gcn_ae_xbd}
\resizebox{\textwidth}{!}{%
\begin{tabular}{c|cccccc|cccccc}
\textbf{Set} & \multicolumn{6}{c|}{\textbf{AE}} & \multicolumn{6}{c}{\textbf{BLDNet}} \\ \hline
\multirow{5}{*}{\textbf{Pinery Bushfire}} &  & \textbf{Acc} & \textbf{Precision} & \textbf{Recall} & \textbf{Specificity} & \textbf{F1} & \textbf{} & \textbf{Acc} & \textbf{Precision} & \textbf{Recall} & \textbf{Specificity} & \textbf{F1} \\
 & \textbf{Train} & 0.6044 & 0.4935 & 0.4811 & 0.7386 & 0.4495 & \textbf{Train} & \textit{\textbf{0.9524}} & \textit{\textbf{0.8444}} & \textit{\textbf{0.9813}} & \textit{\textbf{0.9822}} & \textit{\textbf{0.9023}} \\
 & \textbf{Test} & 0.5000 & 0.3338 & 0.3288 & 0.6357 & 0.3013 & \textbf{Test} & \textit{\textbf{0.7146}} & \textit{\textbf{0.4515}} & \textit{\textbf{0.5057}} & \textit{\textbf{0.7791}} & \textit{\textbf{0.4660}} \\
 & \textbf{Hold} & 0.4773 & 0.3439 & 0.3090 & 0.6651 & 0.3044 & \textbf{Hold} & \textit{\textbf{0.7145}} & \textit{\textbf{0.4517}} & \textit{\textbf{0.4941}} & \textit{\textbf{0.7743}} & \textit{\textbf{0.4646}} \\
 & \textbf{Full} & 0.3909 & 0.2993 & 0.3226 & 0.6028 & 0.2588 & \textbf{Full} & \textit{\textbf{0.7247}} & \textit{\textbf{0.4671}} & \textit{\textbf{0.5182}} & \textit{\textbf{0.7918}} & \textit{\textbf{0.4826}} \\ \hline
\multirow{5}{*}{\textbf{Joplin Tornado}} &  & \textbf{Acc} & \textbf{Precision} & \textbf{Recall} & \textbf{Specificity} & \textbf{F1} & \textbf{} & \textbf{Acc} & \textbf{Precision} & \textbf{Recall} & \textbf{Specificity} & \textbf{F1} \\
 & \textbf{Train} & 0.6337 & 0.5450 & 0.4925 & 0.8033 & 0.5067 & \textbf{Train} & \textit{\textbf{0.9634}} & \textit{\textbf{0.9462}} & \textit{\textbf{0.9696}} & \textit{\textbf{0.9839}} & \textit{\textbf{0.6562}} \\
 & \textbf{Test} & 0.5171 & 0.5188 & 0.4274 & 0.7484 & 0.4457 & \textbf{Test} & \textit{\textbf{0.7610}} & \textit{\textbf{0.6968}} & \textit{\textbf{0.6968}} & \textit{\textbf{0.8802}} & \textit{\textbf{0.6900}} \\
 & \textbf{Hold} & 0.6179 & 0.5382 & 0.4967 & 0.7837 & 0.4967 & \textbf{Hold} & \textit{\textbf{0.7540}} & \textit{\textbf{0.6987}} & \textit{\textbf{0.6992}} & \textit{\textbf{0.8778}} & \textit{\textbf{0.6914}} \\
 & \textbf{Full} & 0.5988 & 0.5400 & 0.4893 & 0.7809 & 0.4961 & \textbf{Full} & \textit{\textbf{0.7606}} & \textit{\textbf{0.7049}} & \textit{\textbf{0.7027}} & \textit{\textbf{0.8810}} & \textit{\textbf{0.6954}} \\ \hline
\multirow{5}{*}{\textbf{Nepal Flooding}} &  & \textbf{Acc} & \textbf{Precision} & \textbf{Recall} & \textbf{Specificity} & \textbf{F1} & \textbf{} & \textbf{Acc} & \textbf{Precision} & \textbf{Recall} & \textbf{Specificity} & \textbf{F1} \\
 & \textbf{Train} & 0.6484 & 0.3976 & 0.6864 & 0.6836 & 0.3905 & \textbf{Train} & \textit{\textbf{0.9707}} & \textit{\textbf{0.9310}} & \textit{\textbf{0.9871}} & \textit{\textbf{0.9889}} & \textit{\textbf{0.9566}} \\
 & \textbf{Test} & 0.5463 & 0.3664 & 0.3701 & 0.6958 & 0.3643 & \textbf{Test} & \textit{\textbf{0.6780}} & \textit{\textbf{0.4907}} & \textit{\textbf{0.4535}} & \textit{\textbf{0.7344}} & \textit{\textbf{0.4623}} \\
 & \textbf{Hold} & 0.5930 & 0.4056 & 0.4121 & 0.7059 & 0.4082 & \textbf{Hold} & \textit{\textbf{0.7086}} & \textit{\textbf{0.5589}} & \textit{\textbf{0.5199}} & \textit{\textbf{0.7613}} & \textit{\textbf{0.5328}} \\
 & \textbf{Full} & 0.5754 & 0.3712 & 0.3733 & 0.6907 & 0.3720 & \textbf{Full} & \textit{\textbf{0.7152}} & \textit{\textbf{0.5585}} & \textit{\textbf{0.5244}} & \textit{\textbf{0.7710}} & \textit{\textbf{0.5363}}
\end{tabular}%
}
\end{table}

\begin{figure}[htb]
    \centering
    \includegraphics[scale=0.5]{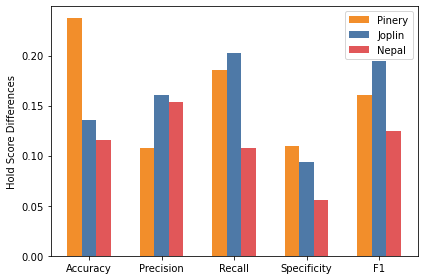}
    \caption{Performance difference with respect to the autoencoder for all three disasters.}
    \label{fig:gcn_diff}
\end{figure}

BLDNet systematically outperforms the autoencoder. Figure \ref{fig:gcn_diff} shows the difference between the BLDNet and autoencoder hold scores for all three disasters. We achieved an average increase of $16.3\%$, $14.05\%$, $16.51\%$, $8.62\%$ and $15.98\%$ for accuracy, precision, recall, specificity and F1 score respectively on the hold set across disasters. Recall saw the highest average improvement which signifies a notable amelioration in the ability to detect the positive class (damage). We also see a trend where specificity tends to be higher than other metrics with a mean difference of $17.95\%$ with the recall. This measure represents the model performance on the negative class as opposed to recall. The difference is more pronounced in highly unbalanced datasets. We computed the Shannon equitability index \cite{shannon_1948} (range 0 to 1 with 1 being balanced and 0 unbalanced) for each of the sets we used in both their original and pruned versions. We also calculated the difference between the specificity and recall scores on the full set for each disaster (Table \ref{tab:shannon-xbd}). First, all pruned sets have a higher index than their original counterparts which means that our data pruning had a positive impact on the unbalanced state of the dataset. Also, the difference between specificity and recall is consistently reduced with lower class imbalance (higher Shannon index). We therefore conclude that decreasing class imbalance and the prevalence of non-damaged samples increased the model's ability to accurately detect damage.

\begin{table}[htb]
\centering
\caption{Shannon Equitability Index for the Chosen xBD Sets and the Difference between Specificity and Recall.}
\label{tab:shannon-xbd}
\resizebox{!}{!}{%
\begin{tabular}{c|c|c|c|}
\cline{2-4}
 & Joplin Tornado & Nepal Flooding & Pinery Bushfire \\ \hline
\multicolumn{1}{|c|}{Shannon equitability original} & 0.813 & 0.558 & 0.247 \\ \hline
\multicolumn{1}{|c|}{Shannon equitability pruned} & 0.878 & 0.597 & 0.495 \\ \hline
\multicolumn{1}{|c|}{Specificity - Recall} & 0.1783 & 0.2466 & 0.2736 \\ \hline
\end{tabular}%
}
\end{table}

\begin{figure}[htb]
    \centering
    \includegraphics[scale=0.6]{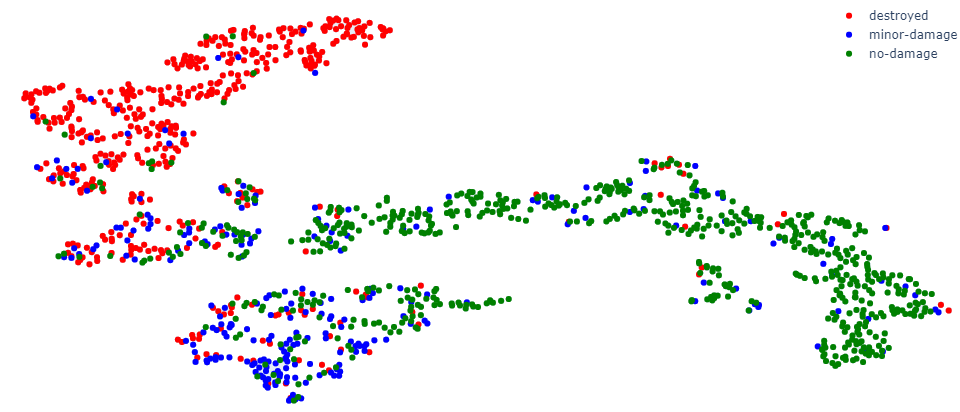}
    \caption{BLDNet node embeddings for Joplin tornado.}
    \label{fig:joplin-tsne}
\end{figure}

Additionally, we investigate the class separation ability of our model by producing a TSNE (t-distributed stochastic neighbor embedding) \cite{van2008visualizing} visualization of the node embeddings produced by BLDNet (Figure \ref{fig:joplin-tsne}). We notice that each class forms a cluster. However, the "minor damage" class lies between the two other classes and does not form a cluster as separated and as condensed as the other two. We see "minor damage" points distributed within the other class clusters. This gives credit to our and Xu et al.'s assumption that intermediate damage classes are ambiguous and harder to distinguish from other classes \cite{xu_building_2019}. These results are achieved without using an ordinal cross entropy loss function and therefore the model had no incentive to treat the classes as being ordered.

\begin{figure}[htb]
    \centering
    \includegraphics[scale=0.5]{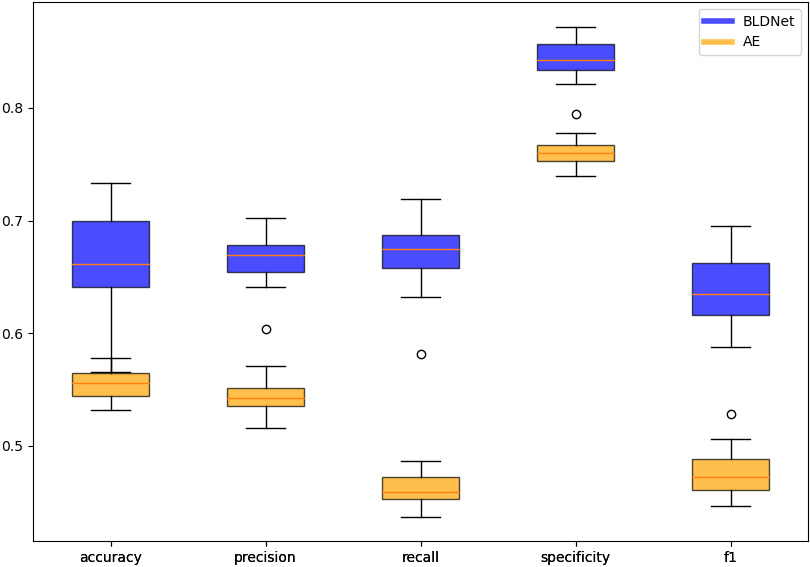}
    \caption{Distribution of the performance metrics obtained after 30 runs with different data samples.}
    \label{fig:ttest_box}
\end{figure}

To verify that the difference in performance between the autoencoder and our model is not situational, we ran the Paired Student t-test \cite{student1908probable} and Wilcoxon signed rank test \cite{wilcoxon1945}. Each of the two models is trained thirty times on a differently sampled training set and its performance is reported on a hold set. To ensure that the independence condition is not violated and that there is no overlap between the re-sampled training sets and the hold set, the latter is separately partitioned and the remainder of the data is randomly sampled into a different training set every run. The variable considered for the test is the difference between the two models' metrics. A separate test is carried for each of the calculated metrics with a significance level of 5\%. The null hypothesis was rejected for every metric, which means that the difference in performance is not due to chance. Looking at the distributions of the metric populations (Figure \ref{fig:ttest_box}), it can be argued that the difference is clearly significant without the need to run a statistical test.

\subsection{Semi-supervised GCN with the Beirut Explosion dataset}
This experiment demonstrates the effectiveness of our method on a different scenario while including meta-features.

We run the experiment with the same model and training configurations that were adopted for the xBD dataset. No data pruning strategy was needed since the area of interest was selected during preprocessing (Figure \ref{fig:beirut_area}). The first experiment was performed without any meta-features and the second with urban domain knowledge injection. Table \ref{tab:beirut} shows the results of both experiments.

\begin{table}[htb]
\centering
\caption{Comparison between BLDNet with or without Meta-features on the Beirut Data, with bold italics for best metrics.}
\label{tab:beirut}
\resizebox{!}{!}{%
\begin{tabular}{c|cccccc}
\multirow{5}{*}{\textbf{Beirut}} &  & \textbf{Accuracy} & \textbf{Precision} & \textbf{Recall} & \textbf{Specificity} & \textbf{F1} \\
 & \textbf{Train} & 0.9534 & 0.8474 & 0.9804 & 0.9817 & 0.9033 \\
 & \textbf{Test} & 0.7415 & \textit{\textbf{0.5930}} & \textit{\textbf{0.6744}} & 0.7810 & \textit{\textbf{0.6189}} \\
 & \textbf{Hold} & 0.7291 & 0.5993 & 0.6826 & 0.7815 & 0.6274 \\
 & \textbf{Full} & 0.7460 & 0.6241 & 0.7276 & 0.8031 & 0.6595 \\ \hline
\multirow{5}{*}{\textbf{Beirut Meta-features}} & \textbf{} & \textbf{Accuracy} & \textbf{Precision} & \textbf{Recall} & \textbf{Specificity} & \textbf{F1} \\
 & \textbf{Train} & \textit{\textbf{0.9963}} & \textit{\textbf{0.9926}} & \textit{\textbf{0.9985}} & \textit{\textbf{0.9985}} & \textit{\textbf{0.9955}} \\
 & \textbf{Test} & \textit{\textbf{0.7878}} & 0.5779 & 0.6571 & \textit{\textbf{0.7827}} & 0.6100 \\
 & \textbf{Hold} & \textit{\textbf{0.8170}} & \textit{\textbf{0.6533}} & \textit{\textbf{0.7329}} & \textit{\textbf{0.8124}} & \textit{\textbf{0.6866}} \\
 & \textbf{Full} & \textit{\textbf{0.8133}} & \textit{\textbf{0.6409}} & \textit{\textbf{0.7407}} & \textit{\textbf{0.8178}} & \textit{\textbf{0.6808}}
\end{tabular}%
}
\end{table}

In most aspects, injection of meta-features yielded better results. On the hold set, we achieve an average increment of $5.65\%$ over the five metrics. The least improvement is $3.09\%$ for specificity and the highest is $8.79\%$ for accuracy. This shows that augmenting the images with additional contextual features helps the model better estimate how each building was impacted by the disaster. This improvement was achieved despite $32.18\%$ of the data being missing on average across the meta-features. We believe that domain knowledge injection can add even more value with better conditioned data and with the possibility of embedding meta-features at the edge level and not just the node level.


\section{Conclusion}\label{sec:conclusion}
Disaster informatics is essential for planning effective responses. With the availability of satellite images and remote sensing, artificial intelligence can be valuable for urban data science and disaster planning.

Given that the majority of the destruction is usually clustered around the epicenter of the disaster, it can be expected that neighboring buildings would have a similar damage signature. However, in many cases the urban layout of cities is non-uniform and it is very common to observe neighboring buildings of different age, height and function. This adds an additional layer of interactions that makes these buildings sustain damage differently. 

In this work, we have presented a graph formulation that allows representing damage data while connecting buildings. We had proposed BLDNet, a semi-supervised GCN with a Siamese CNN backbone for extracting local features and aggregating them with neighboring features. Through semi-supervised learning, few labeled images enabled BLDNet to obtain predictions in an efficient manner. We demonstrated the effectiveness of BLDNet on the xBD dataset and compared it to the multiresolution autoencoder. We also showcased BLDNet's effectiveness on the 2020 Beirut Port explosion where augmenting the damage images with contextual building meta-features improved model performance.

Based on the insights gathered throughout this work, potential directions for future research include using additional satellite imagery bands in addition to color bands and applying our graph formulation on geotagged street view images of damaged buildings.

\section*{Acknowledgments}
The authors would like to thank Maxar Technologies for donating satellite images of the Beirut area during dates surrounding the August 4 explosion. We also credit the Open Map Lebanon team for promptly providing the building damage annotations for the Beirut Port explosion.

\bibliographystyle{unsrt}  
\bibliography{references}

\begin{thebibliography}{10}

\bibitem{khan_forest_2017}
Salman~H. Khan, Xuming He, Fatih Porikli, and Mohammed Bennamoun.
\newblock Forest {Change} {Detection} in {Incomplete} {Satellite} {Images}
  {With} {Deep} {Neural} {Networks}.
\newblock {\em IEEE Transactions on Geoscience and Remote Sensing},
  55(9):5407--5423, September 2017.

\bibitem{tan_improved_2013}
Bin Tan, Jeffrey~G. Masek, Robert Wolfe, Feng Gao, Chengquan Huang, Eric~F.
  Vermote, Joseph~O. Sexton, and Greg Ederer.
\newblock Improved forest change detection with terrain illumination corrected
  {Landsat} images.
\newblock {\em Remote Sensing of Environment}, 136:469--483, September 2013.

\bibitem{woodcock_monitoring_2001}
Curtis~E Woodcock, Scott~A Macomber, Mary Pax-Lenney, and Warren~B Cohen.
\newblock Monitoring large areas for forest change using {Landsat}:
  {Generalization} across space, time and {Landsat} sensors.
\newblock {\em Remote Sensing of Environment}, 78(1-2):194--203, October 2001.

\bibitem{rokni_new_2015}
Komeil Rokni, Anuar Ahmad, Karim Solaimani, and Sharifeh Hazini.
\newblock A new approach for surface water change detection: {Integration} of
  pixel level image fusion and image classification techniques.
\newblock {\em International Journal of Applied Earth Observation and
  Geoinformation}, 34:226--234, February 2015.

\bibitem{song_change_2019}
Ahram Song, Yeji Kim, and Yongil Kim.
\newblock Change {Detection} of {Surface} {Water} in {Remote} {Sensing}
  {Images} {Based} on {Fully} {Convolutional} {Network}.
\newblock {\em Journal of Coastal Research}, 91(sp1):426, August 2019.

\bibitem{gao_sea_2019}
Feng Gao, Xiao Wang, Yunhao Gao, Junyu Dong, and Shengke Wang.
\newblock Sea {Ice} {Change} {Detection} in {SAR} {Images} {Based} on
  {Convolutional}-{Wavelet} {Neural} {Networks}.
\newblock {\em IEEE Geoscience and Remote Sensing Letters}, 16(8):1240--1244,
  August 2019.

\bibitem{gao_transferred_2019}
Yunhao Gao, Feng Gao, Junyu Dong, and Shengke Wang.
\newblock Transferred {Deep} {Learning} for {Sea} {Ice} {Change} {Detection}
  {From} {Synthetic}-{Aperture} {Radar} {Images}.
\newblock {\em IEEE Geoscience and Remote Sensing Letters}, 16(10):1655--1659,
  October 2019.

\bibitem{ding_automatic_2016}
Anzi Ding, Qingyong Zhang, Xinmin Zhou, and Bicheng Dai.
\newblock Automatic recognition of landslide based on {CNN} and texture change
  detection.
\newblock In {\em 2016 31st {Youth} {Academic} {Annual} {Conference} of
  {Chinese} {Association} of {Automation} ({YAC})}, pages 444--448, Wuhan,
  Hubei Province, China, November 2016. IEEE.

\bibitem{bovolo_split-based_2007}
Francesca Bovolo and Lorenzo Bruzzone.
\newblock A {Split}-{Based} {Approach} to {Unsupervised} {Change} {Detection}
  in {Large}-{Size} {Multitemporal} {Images}: {Application} to
  {Tsunami}-{Damage} {Assessment}.
\newblock {\em IEEE Transactions on Geoscience and Remote Sensing},
  45(6):1658--1670, June 2007.

\bibitem{sublime_automatic_2019}
Jérémie Sublime and Ekaterina Kalinicheva.
\newblock Automatic {Post}-{Disaster} {Damage} {Mapping} {Using}
  {Deep}-{Learning} {Techniques} for {Change} {Detection}: {Case} {Study} of
  the {Tohoku} {Tsunami}.
\newblock {\em Remote Sensing}, 11(9):1123, May 2019.

\bibitem{fraser_multitemporal_2003}
R.~H. Fraser, R.~Fernandes, and R.~Latifovic.
\newblock Multi‐temporal {Mapping} of {Burned} {Forest} over {Canada} {Using}
  {Satellite}‐based {Change} {Metrics}.
\newblock {\em Geocarto International}, 18(2):37--47, June 2003.

\bibitem{zhang_detecting_2019}
Chi Zhang, Shiqing Wei, Shunping Ji, and Meng Lu.
\newblock Detecting {Large}-{Scale} {Urban} {Land} {Cover} {Changes} from
  {Very} {High} {Resolution} {Remote} {Sensing} {Images} {Using} {CNN}-{Based}
  {Classification}.
\newblock {\em ISPRS International Journal of Geo-Information}, 8(4):189, April
  2019.

\bibitem{cao_land-use_2019}
Cong Cao, Suzana Dragićević, and Songnian Li.
\newblock Land-{Use} {Change} {Detection} with {Convolutional} {Neural}
  {Network} {Methods}.
\newblock {\em Environments}, 6(2):25, February 2019.

\bibitem{lu_detection_2010}
Dengsheng Lu.
\newblock Detection of urban expansion in an urban-rural landscape with
  multitemporal {QuickBird} images.
\newblock {\em Journal of Applied Remote Sensing}, 4(1):041880, September 2010.

\bibitem{kerner_toward_2019}
Hannah~Rae Kerner, Kiri~L. Wagstaff, Brian~D. Bue, Patrick~C. Gray, James~F.
  Bell, and Heni Ben~Amor.
\newblock Toward {Generalized} {Change} {Detection} on {Planetary} {Surfaces}
  {With} {Convolutional} {Autoencoders} and {Transfer} {Learning}.
\newblock {\em IEEE Journal of Selected Topics in Applied Earth Observations
  and Remote Sensing}, 12(10):3900--3918, October 2019.

\bibitem{robila_use_2006}
Stefan~A. Robila.
\newblock Use of {Remote} {Sensing} {Applications} and its {Implications} to
  the {Society}.
\newblock In {\em 2006 {IEEE} {International} {Symposium} on {Technology} and
  {Society}}, pages 1--6, Queens, NY, USA, June 2006. IEEE.

\bibitem{gupta_xbd_2019}
Ritwik Gupta, Richard Hosfelt, Sandra Sajeev, Nirav Patel, Bryce Goodman, Jigar
  Doshi, Eric Heim, Howie Choset, and Matthew Gaston.
\newblock {xBD}: {A} {Dataset} for {Assessing} {Building} {Damage} from
  {Satellite} {Imagery}.
\newblock {\em arXiv:1911.09296 [cs]}, November 2019.

\bibitem{asokan_change_2019}
Anju Asokan and J.~Anitha.
\newblock Change detection techniques for remote sensing applications: a
  survey.
\newblock {\em Earth Science Informatics}, 12(2):143--160, June 2019.

\bibitem{shi_change_2020}
Wenzhong Shi, Min Zhang, Rui Zhang, Shanxiong Chen, and Zhao Zhan.
\newblock Change {Detection} {Based} on {Artificial} {Intelligence}:
  {State}-of-the-{Art} and {Challenges}.
\newblock {\em Remote Sensing}, 12(10):1688, May 2020.

\bibitem{khelifi_deep_2020}
Lazhar Khelifi and Max Mignotte.
\newblock Deep {Learning} for {Change} {Detection} in {Remote} {Sensing}
  {Images}: {Comprehensive} {Review} and {Meta}-{Analysis}.
\newblock {\em arXiv:2006.05612 [cs]}, June 2020.

\bibitem{peng_end--end_2019}
Daifeng Peng, Yongjun Zhang, and Haiyan Guan.
\newblock End-to-{End} {Change} {Detection} for {High} {Resolution} {Satellite}
  {Images} {Using} {Improved} {UNet}++.
\newblock {\em Remote Sensing}, 11(11):1382, June 2019.

\bibitem{zhan_change_2017}
Yang Zhan, Kun Fu, Menglong Yan, Xian Sun, Hongqi Wang, and Xiaosong Qiu.
\newblock Change {Detection} {Based} on {Deep} {Siamese} {Convolutional}
  {Network} for {Optical} {Aerial} {Images}.
\newblock {\em IEEE Geoscience and Remote Sensing Letters}, 14(10):1845--1849,
  October 2017.

\bibitem{daudt_urban_2018}
Rodrigo~Caye Daudt, Bertr Le~Saux, Alexandre Boulch, and Yann Gousseau.
\newblock Urban {Change} {Detection} for {Multispectral} {Earth} {Observation}
  {Using} {Convolutional} {Neural} {Networks}.
\newblock In {\em {IGARSS} 2018 - 2018 {IEEE} {International} {Geoscience} and
  {Remote} {Sensing} {Symposium}}, pages 2115--2118, Valencia, July 2018. IEEE.

\bibitem{lyu_learning_2016}
Haobo Lyu, Hui Lu, and Lichao Mou.
\newblock Learning a {Transferable} {Change} {Rule} from a {Recurrent} {Neural}
  {Network} for {Land} {Cover} {Change} {Detection}.
\newblock {\em Remote Sensing}, 8(6):506, June 2016.

\bibitem{mou_learning_2019}
Lichao Mou, Lorenzo Bruzzone, and Xiao~Xiang Zhu.
\newblock Learning {Spectral}-{Spatial}-{Temporal} {Features} via a {Recurrent}
  {Convolutional} {Neural} {Network} for {Change} {Detection} in
  {Multispectral} {Imagery}.
\newblock {\em IEEE Transactions on Geoscience and Remote Sensing},
  57(2):924--935, February 2019.

\bibitem{chen_change_2020}
Hongruixuan Chen, Chen Wu, Bo~Du, Liangpei Zhang, and Le~Wang.
\newblock Change {Detection} in {Multisource} {VHR} {Images} via {Deep}
  {Siamese} {Convolutional} {Multiple}-{Layers} {Recurrent} {Neural} {Network}.
\newblock {\em IEEE Transactions on Geoscience and Remote Sensing},
  58(4):2848--2864, April 2020.

\bibitem{jing_object-based_2020}
Ran Jing, Shuang Liu, Zhaoning Gong, Zhiheng Wang, Hongliang Guan, Atul Gautam,
  and Wenji Zhao.
\newblock Object-based change detection for {VHR} remote sensing images based
  on a {Trisiamese}-{LSTM}.
\newblock {\em International Journal of Remote Sensing}, 41(16):6209--6231,
  August 2020.

\bibitem{ji_building_2019}
Shunping Ji, Yanyun Shen, Meng Lu, and Yongjun Zhang.
\newblock Building {Instance} {Change} {Detection} from {Large}-{Scale}
  {Aerial} {Images} using {Convolutional} {Neural} {Networks} and {Simulated}
  {Samples}.
\newblock {\em Remote Sensing}, 11(11):1343, June 2019.

\bibitem{nex_structural_2019}
Francesco Nex, Diogo Duarte, Fabio~Giulio Tonolo, and Norman Kerle.
\newblock Structural {Building} {Damage} {Detection} with {Deep} {Learning}:
  {Assessment} of a {State}-of-the-{Art} {CNN} in {Operational} {Conditions}.
\newblock {\em Remote Sensing}, 11(23):2765, November 2019.

\bibitem{jiang_pga-siamnet_2020}
Huiwei Jiang, Xiangyun Hu, Kun Li, Jinming Zhang, Jinqi Gong, and Mi~Zhang.
\newblock {PGA}-{SiamNet}: {Pyramid} {Feature}-{Based} {Attention}-{Guided}
  {Siamese} {Network} for {Remote} {Sensing} {Orthoimagery} {Building} {Change}
  {Detection}.
\newblock {\em Remote Sensing}, 12(3):484, February 2020.

\bibitem{kalantar_assessment_2020}
Bahareh Kalantar, Naonori Ueda, Husam A.~H. Al-Najjar, and Alfian~Abdul Halin.
\newblock Assessment of {Convolutional} {Neural} {Network} {Architectures} for
  {Earthquake}-{Induced} {Building} {Damage} {Detection} based on {Pre}- and
  {Post}-{Event} {Orthophoto} {Images}.
\newblock {\em Remote Sensing}, 12(21):3529, October 2020.

\bibitem{miura_deep_2020}
Hiroyuki Miura, Tomohiro Aridome, and Masashi Matsuoka.
\newblock Deep {Learning}-{Based} {Identification} of {Collapsed},
  {Non}-{Collapsed} and {Blue} {Tarp}-{Covered} {Buildings} from
  {Post}-{Disaster} {Aerial} {Images}.
\newblock {\em Remote Sensing}, 12(12):1924, June 2020.

\bibitem{wheeler_deep_2020}
Bradley~J. Wheeler and Hassan~A. Karimi.
\newblock Deep {Learning}-{Enabled} {Semantic} {Inference} of {Individual}
  {Building} {Damage} {Magnitude} from {Satellite} {Images}.
\newblock {\em Algorithms}, 13(8):195, August 2020.

\bibitem{kipf_semi-supervised_2017}
Thomas~N. Kipf and Max Welling.
\newblock Semi-{Supervised} {Classification} with {Graph} {Convolutional}
  {Networks}.
\newblock {\em arXiv:1609.02907 [cs, stat]}, February 2017.

\bibitem{hong_graph_2020}
Danfeng Hong, Lianru Gao, Jing Yao, Bing Zhang, Antonio Plaza, and Jocelyn
  Chanussot.
\newblock Graph {Convolutional} {Networks} for {Hyperspectral} {Image}
  {Classification}.
\newblock {\em IEEE Transactions on Geoscience and Remote Sensing}, pages
  1--13, 2020.

\bibitem{chaudhuri_siamese_2019}
Ushasi Chaudhuri, Biplab Banerjee, and Avik Bhattacharya.
\newblock Siamese graph convolutional network for content based remote sensing
  image retrieval.
\newblock {\em Computer Vision and Image Understanding}, 184:22--30, July 2019.

\bibitem{saha_semisupervised_2020}
Sudipan Saha, Francesca Bovolo, and Lorenzo Bruzzone.
\newblock Semisupervised {Change} {Detection} {Using} {Graph} {Convolutional}
  {Network}.
\newblock {\em IEEE Geoscience and Remote Sensing Letters}, 18(4):607 -- 611,
  2020.

\bibitem{wang_change_2018}
Qing Wang, Xiaodong Zhang, Guanzhou Chen, Fan Dai, Yuanfu Gong, and Kun Zhu.
\newblock Change detection based on {Faster} {R}-{CNN} for high-resolution
  remote sensing images.
\newblock {\em Remote Sensing Letters}, 9(10):923--932, October 2018.

\bibitem{weber_building_2020}
Ethan Weber and Hassan Kané.
\newblock Building {Disaster} {Damage} {Assessment} in {Satellite} {Imagery}
  with {Multi}-{Temporal} {Fusion}.
\newblock {\em arXiv:2004.05525 [cs]}, April 2020.

\bibitem{su_technical_2020}
Jinhua Su, Yanbing Bai, Xingrui Wang, Dong Lu, Bo~Zhao, Hanfang Yang, Erick
  Mas, and Shunichi Koshimura.
\newblock Technical {Solution} {Discussion} for {Key} {Challenges} of
  {Operational} {Convolutional} {Neural} {Network}-{Based} {Building}-{Damage}
  {Assessment} from {Satellite} {Imagery}: {Perspective} from {Benchmark} {xBD}
  {Dataset}.
\newblock {\em Remote Sensing}, 12(22):3808, November 2020.

\bibitem{xu_building_2019}
Joseph~Z. Xu, Wenhan Lu, Zebo Li, Pranav Khaitan, and Valeriya Zaytseva.
\newblock Building {Damage} {Detection} in {Satellite} {Imagery} {Using}
  {Convolutional} {Neural} {Networks}.
\newblock {\em arXiv:1910.06444 [cs, eess, stat]}, October 2019.

\bibitem{benson_assessing_2020}
Vitus Benson and Alexander Ecker.
\newblock Assessing out-of-domain generalization for robust building damage
  detection.
\newblock {\em arXiv:2011.10328 [cs]}, November 2020.

\bibitem{bai_pyramid_2020}
Yanbing Bai, Junjie Hu, Jinhua Su, Xing Liu, Haoyu Liu, Xianwen He, Shengwang
  Meng, Erick Mas, and Shunichi Koshimura.
\newblock Pyramid {Pooling} {Module}-{Based} {Semi}-{Siamese} {Network}: {A}
  {Benchmark} {Model} for {Assessing} {Building} {Damage} from {xBD}
  {Satellite} {Imagery} {Datasets}.
\newblock {\em Remote Sensing}, 12(24):4055, December 2020.

\bibitem{yang_transferability_2021}
Wanting Yang, Xianfeng Zhang, and Peng Luo.
\newblock Transferability of {Convolutional} {Neural} {Network} {Models} for
  {Identifying} {Damaged} {Buildings} {Due} to {Earthquake}.
\newblock {\em Remote Sensing}, 13(3):504, January 2021.

\bibitem{zheng_building_2021}
Zhuo Zheng, Yanfei Zhong, Junjue Wang, Ailong Ma, and Liangpei Zhang.
\newblock Building damage assessment for rapid disaster response with a deep
  object-based semantic change detection framework: {From} natural disasters to
  man-made disasters.
\newblock {\em Remote Sensing of Environment}, 265:112636, November 2021.

\bibitem{pati_novel_2020}
Chinmayee Pati, Ashok~K. Panda, Ajaya~Kumar Tripathy, Sateesh~K. Pradhan, and
  Srikanta Patnaik.
\newblock A novel hybrid machine learning approach for change detection in
  remote sensing images.
\newblock {\em Engineering Science and Technology, an International Journal},
  23(5):973--981, October 2020.

\bibitem{tilon_post-disaster_2020}
Sofia Tilon, Francesco Nex, Norman Kerle, and George Vosselman.
\newblock Post-{Disaster} {Building} {Damage} {Detection} from {Earth}
  {Observation} {Imagery} {Using} {Unsupervised} and {Transferable} {Anomaly}
  {Detecting} {Generative} {Adversarial} {Networks}.
\newblock {\em Remote Sensing}, 12(24):4193, December 2020.

\bibitem{peng_semicdnet_2021}
Daifeng Peng, Lorenzo Bruzzone, Yongjun Zhang, Haiyan Guan, Haiyong Ding, and
  Xu~Huang.
\newblock {SemiCDNet}: {A} {Semisupervised} {Convolutional} {Neural} {Network}
  for {Change} {Detection} in {High} {Resolution} {Remote}-{Sensing} {Images}.
\newblock {\em IEEE Transactions on Geoscience and Remote Sensing},
  59(7):5891--5906, July 2021.

\bibitem{khan_graph_2019}
Nagma Khan, Ushasi Chaudhuri, Biplab Banerjee, and Subhasis Chaudhuri.
\newblock Graph convolutional network for multi-label {VHR} remote sensing
  scene recognition.
\newblock {\em Neurocomputing}, 357:36--46, September 2019.

\bibitem{chen_multi-label_2019}
Zhao-Min Chen, Xiu-Shen Wei, Peng Wang, and Yanwen Guo.
\newblock Multi-{Label} {Image} {Recognition} {With} {Graph} {Convolutional}
  {Networks}.
\newblock In {\em 2019 {IEEE}/{CVF} {Conference} on {Computer} {Vision} and
  {Pattern} {Recognition} ({CVPR})}, pages 5172--5181, Long Beach, CA, USA,
  June 2019. IEEE.

\bibitem{mou_nonlocal_2020}
Lichao Mou, Xiaoqiang Lu, Xuelong Li, and Xiao~Xiang Zhu.
\newblock Nonlocal {Graph} {Convolutional} {Networks} for {Hyperspectral}
  {Image} {Classification}.
\newblock {\em IEEE Transactions on Geoscience and Remote Sensing}, pages
  1--12, 2020.

\bibitem{lee_two_1980}
D.~T. Lee and B.~J. Schachter.
\newblock Two algorithms for constructing a {Delaunay} triangulation.
\newblock {\em International Journal of Computer \& Information Sciences},
  9(3):219--242, June 1980.

\bibitem{he2015deep}
Kaiming He, Xiangyu Zhang, Shaoqing Ren, and Jian Sun.
\newblock Deep residual learning for image recognition, 2015.

\bibitem{calderisi_improve_2019}
Marco Calderisi, Gabriele Galatolo, Ilaria Ceppa, Tommaso Motta, and Francesco
  Vergentini.
\newblock Improve {Image} {Classification} {Tasks} {Using} {Simple}
  {Convolutional} {Architectures} with {Processed} {Metadata} {Injection}.
\newblock In {\em 2019 {IEEE} {Second} {International} {Conference} on
  {Artificial} {Intelligence} and {Knowledge} {Engineering} ({AIKE})}, pages
  223--230, Sardinia, Italy, June 2019. IEEE.

\bibitem{ellen_improving_2019}
Jeffrey~S. Ellen, Casey~A. Graff, and Mark~D. Ohman.
\newblock Improving plankton image classification using context metadata.
\newblock {\em Limnology and Oceanography: Methods}, 17(8):439--461, August
  2019.

\bibitem{bucknell_2018}
Alice Bucknell.
\newblock High contrast: We chart beirut's ever-changing architecture scene,
  Aug 2018.

\bibitem{reliefweb}
Lebanon: Beirut port explosions - aug 2020.

\bibitem{rigby_preliminary_2020}
S.~E. Rigby, T.~J. Lodge, S.~Alotaibi, A.~D. Barr, S.~D. Clarke, G.~S. Langdon,
  and A.~Tyas.
\newblock Preliminary yield estimation of the 2020 {Beirut} explosion using
  video footage from social media.
\newblock {\em Shock Waves}, September 2020.

\bibitem{krayem_machine_2021}
Alaa Krayem, Aram Yeretzian, Ghaleb Faour, and Sara Najem.
\newblock Machine learning for buildings’ characterization and power-law
  recovery of urban metrics.
\newblock {\em PLOS ONE}, 16(1):e0246096, January 2021.

\bibitem{fey_fast_2019}
Matthias Fey and Jan~Eric Lenssen.
\newblock Fast {Graph} {Representation} {Learning} with {PyTorch} {Geometric}.
\newblock {\em arXiv:1903.02428 [cs, stat]}, April 2019.

\bibitem{kingma2017adam}
Diederik~P. Kingma and Jimmy Ba.
\newblock Adam: A method for stochastic optimization, 2017.

\bibitem{ienco_semi-supervised_2018}
Dino Ienco and Ruggero~G. Pensa.
\newblock Semi-{Supervised} {Clustering} {With} {Multiresolution}
  {Autoencoders}.
\newblock In {\em 2018 {International} {Joint} {Conference} on {Neural}
  {Networks} ({IJCNN})}, pages 1--8, Rio de Janeiro, July 2018. IEEE.

\bibitem{shannon_1948}
C.~E. Shannon.
\newblock A mathematical theory of communication.
\newblock {\em The Bell System Technical Journal}, 27(3):379--423, 1948.

\bibitem{van2008visualizing}
Laurens Van~der Maaten and Geoffrey Hinton.
\newblock Visualizing data using t-sne.
\newblock {\em Journal of machine learning research}, 9(11), 2008.

\bibitem{student1908probable}
Student.
\newblock The probable error of a mean.
\newblock {\em Biometrika}, pages 1--25, 1908.

\bibitem{wilcoxon1945}
Frank Wilcoxon.
\newblock Individual comparisons by ranking methods.
\newblock {\em Biometrics Bulletin}, 1(6):80--83, 1945.

\end{thebibliography}

\end{document}